\documentclass[11pt,letterpaper]{article}
\usepackage{emnlp2016}
\usepackage{times}
\usepackage{latexsym}
\usepackage{url}
\usepackage{amssymb}
\usepackage{amsmath}
\usepackage{multirow}
\usepackage{graphicx}
\usepackage{floatrow}
\newfloatcommand{capbtabbox}{table}[][0.48\columnwidth] 
\usepackage{array}
\usepackage{bm}
\usepackage[usenames, dvipsnames]{color}
\usepackage{CJKutf8}

\DeclareMathOperator*{\argmax}{arg\,max}
\makeatletter 
\newcommand\figcaption{\def\@captype{figure}\caption} 
\newcommand\tabcaption{\def\@captype{table}\caption} 
\makeatother

\emnlpfinalcopy

\def\emnlppaperid{xxx}

\title{Dataset and Neural Recurrent Sequence Labeling Model for Open-Domain Factoid Question Answering}

\author{Peng Li, Wei Li, Zhengyan He, Xuguang Wang,
	Ying Cao, Jie Zhou, Wei Xu\\
	Baidu Research - Institute of Deep Learning\\
	{\tt \{lipeng17,liwei26,hezhengyan,wangxuguang,caoying03,}\\
	{\tt zhoujie01,wei.xu\}@baidu.com}
}

\date{}

\begin{document}
\begin{CJK}{UTF8}{bsmi}

\maketitle
	
\begin{abstract}
	While question answering (QA) with neural network, i.e. neural QA, has achieved promising results in recent years, lacking of large scale real-word QA dataset is still a challenge for developing and evaluating neural QA system.
	To alleviate this problem, we propose a large scale human annotated real-world QA dataset WebQA with more than 42k questions and 556k evidences.
	As existing neural QA methods resolve QA either as sequence generation or classification/ranking problem, they face challenges of expensive $\mathrm{softmax}$ computation, unseen answers handling or separate candidate answer generation component.
	In this work, we cast neural QA as a sequence labeling problem and propose an end-to-end sequence labeling model, which overcomes all the above challenges.
	Experimental results on WebQA show that our model outperforms the baselines significantly with an F1 score of 74.69\% with word-based input, and the performance drops only 3.72 F1 points with more challenging character-based input.
\end{abstract}

\section{Introduction}
\label{sec:intro}
Question answering (QA) with neural network, i.e. neural QA, is an active research direction along the road towards the long-term AI goal of building general dialogue agents~\cite{weston:2015:arxiv}.  Unlike conventional methods, neural QA does not rely on feature engineering and is (at least nearly) end-to-end trainable. It reduces the requirement for domain specific knowledge significantly and makes domain adaption easier. Therefore, it has attracted intensive attention in recent years.

Resolving QA problem requires several fundamental abilities including reasoning, memorization, etc. 
Various neural methods have been proposed to improve such abilities,
including neural tensor networks~\cite{socher:2013:NIPS}, recursive networks~\cite{iyyer:2014:EMNLP}, convolution neural networks~\cite{yih:2014:ACL,dong:2015:ACL,yin:2016:arxiv},  attention models~\cite{hermann:2015:arxiv,yin:2016:arxiv,santos:2016:arxiv}, and memories~\cite{graves:2014:arixv,weston:2014:arxiv,kumar:2015:arxiv,bordes:2015:arxiv,sukhbaatar:2015:NIPS}, etc.
These methods achieve promising results on various datasets, which demonstrates the high potential of  neural QA. However, we believe there are still two major challenges for neural QA:

{\bf System development and/or evaluation on real-world data:}
Although several high quality and well-designed QA datasets have been proposed in recent years, there are still problems about using them to develop and/or evaluate QA system under real-world settings due to data size and the way they are created.
For example, bAbI~\cite{weston:2015:arxiv} and the 30M Factoid Question-Answer Corpus~\cite{serban:2016:arxiv} are artificially synthesized; 
the TREC datasets~\cite{harman:2006:trec}, Free917~\cite{cai:2013:ACL} and WebQuestions~\cite{berant:2013:EMNLP} are human generated but only have few thousands of questions;
SimpleQuestions~\cite{bordes:2015:arxiv} and the CNN and Daily Mail news datasets~\cite{hermann:2015:arxiv} are large but generated under controlled conditions. 
Thus, a new large-scale real-world QA dataset is needed.

{\bf A new design choice for answer production besides sequence generation and classification/ranking:}
Without loss of generality, the methods used for producing answers in existing neural QA works can be roughly categorized into the sequence generation type and the classification/ranking type.
The former generates answers word by word, e.g.~\cite{weston:2015:arxiv,kumar:2015:arxiv,hermann:2015:arxiv}.
As it generally involves $\mathrm{softmax}$ computation over a large vocabulary, the computational cost is remarkably high and it is hard to produce answers with out-of-vocabulary word.
The latter produces answers by classification over a predefined set of answers, e.g.~\cite{sukhbaatar:2015:NIPS}, or ranking given candidates by model score, e.g.~\cite{yin:2016:arxiv}.
Although it generally has lower computational cost than the former, it either also has difficulties in handling unseen answers or requires an extra candidate generating component which is hard for end-to-end training.
Above all, we need a new design choice for answer production that is both computationally effective and capable of handling unseen words/answers.

In this work, we address the above two challenges by a new dataset and a new neural QA model. Our contributions are two-fold:
\begin{itemize}
	\item We propose a new {\bf large-scale real-world} factoid QA dataset WebQA with more than 42k questions and 566k evidences, where an evidence is a piece of text that contains relevant information to answer the question.
	On one hand, our dataset is an order of magnitude larger than existing real-world QA datasets~\cite{harman:2006:trec,cai:2013:ACL,berant:2013:EMNLP}, which are generally insufficient to train an end-to-end QA system.
	On the other hand, all questions in our dataset are asked by {\em real-world users in daily life}, which is significantly more close to real-world settings than those generated under controlled conditions~\cite{bordes:2015:arxiv,hermann:2015:arxiv}.
	Besides, as we also provide multiple human annotated evidences for each question, the dataset can be used in research such as evidence ranking and answer sentence selection as well.
	\vspace{-0.5em}
	\item We introduce an end-to-end sequence labeling technique into neural QA as a new design choice for answer production.
	Mimicking how humans find answers using search engine, we use conditional random field (CRF)~\cite{lafferty:2001:CRF} to label the answer of a question from retrieved evidence.
	We avoid feature engineering by computing features with a neural model jointly trained with  CRF. As our model does not rely on predefined vocabulary or candidates, it can handle unseen words/answers easily and get rid of expensive $\mathrm{softmax}$ computation.
\end{itemize}
Experimental results show that our model outperforms baselines with a large margin on the WebQA dataset, indicating that it is effective. Furthermore, our model even achieves an F1 score of 70.97\% on character-based input, which is comparable with the 74.69\% F1 score on word-based input, demonstrating that our model is robust.
\begin{figure}
	\centering\footnotesize
	\begin{tabular}{|lp{0.81\linewidth}|}
		\hline
		Q: & Who is the first wife of Albert Einstein ?\\
		E: & Einstein/O married/O his/O first/O wife/O Mileva/B Mari{\' c}/I in/O 1903/O\\
		A: & Mileva Mari{\' c}\\
		\hline
	\end{tabular}
	\caption{Factoid QA as sequence labeling.}
	\label{fig:qa}
\end{figure}
\begin{figure*}
	\centering\small
	\includegraphics[width=0.8\linewidth]{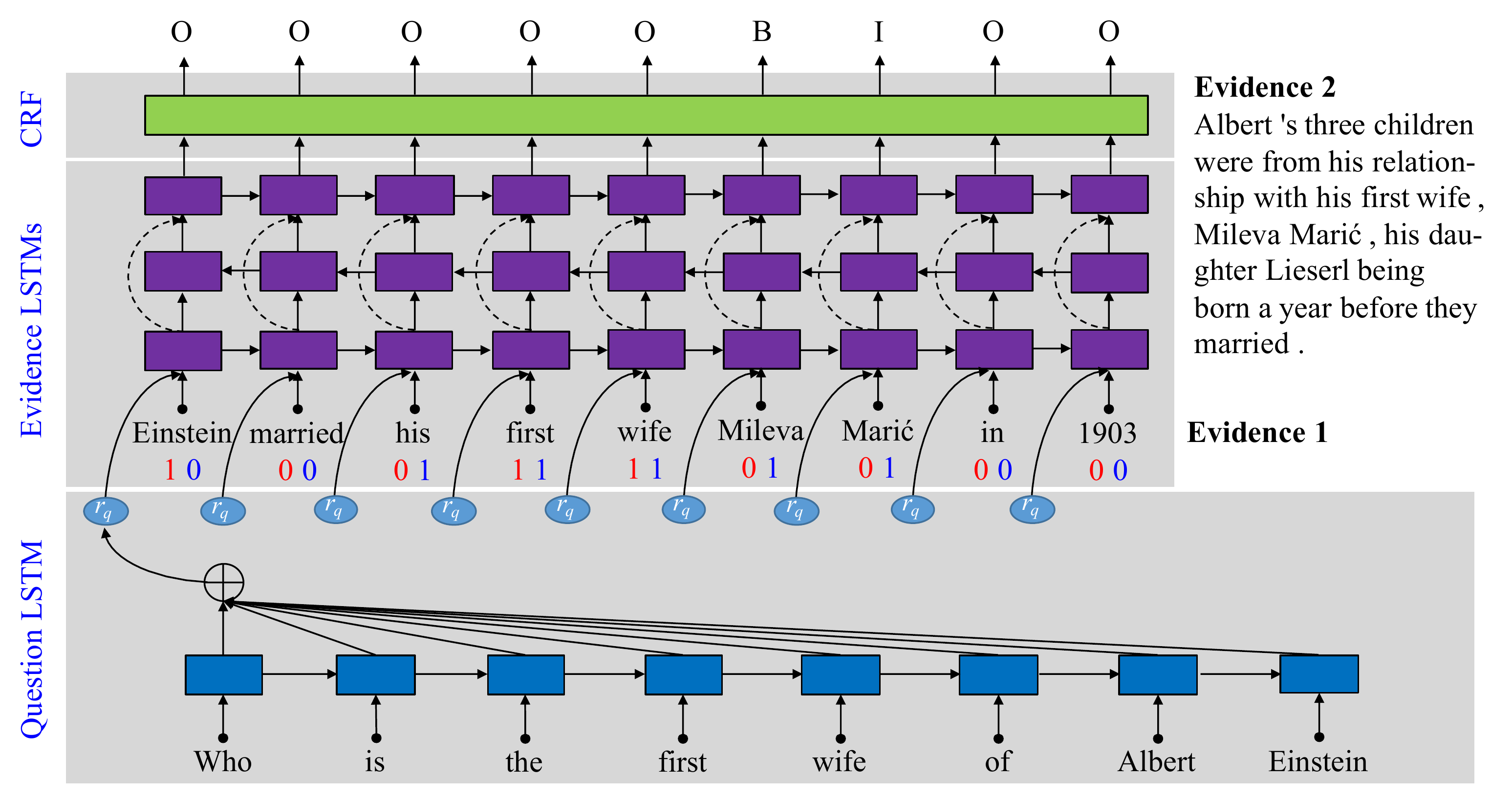}
	\caption{Neural recurrent sequence labeling model for factoid QA. The model consists of three components: ``Question LSTM'' for computing question representation ($r_q$), ``Evidence LSTMs'' for analyzing evidence, and ``CRF'' for producing label sequence which indicates whether each word in the evidence is at the beginning (B), inside (I) or outside (O) of the answer. Each word in the evidence is also equipped with two 0-1 features (see Section~\ref{sec:model:evidence-lstm}). We plot $r_q$ multiple times for clarity.}
	\label{fig:main-figure}
\end{figure*}

\section{Factoid QA as Sequence Labeling}
In this work, we focus on open-domain factoid QA. Taking Figure~\ref{fig:qa} as an example, we formalize the problem as follows: given each question Q, we have one or more evidences E, and the task is to produce the answer A, where an evidence is a piece of text of any length that contains relevant information to answer the question.
The advantage of this formalization is that evidences can be retrieved from web or unstructured knowledge base, which can improve system coverage significantly. 

Inspired by~\cite{yao:2013:NAACL-HLT}, we introduce end-to-end sequence labeling as a new design choice for answer production in neural QA.
Given a question and an evidence, we use CRF~\cite{lafferty:2001:CRF} to assign a label to each word in the evidence to indicate whether the word is at the beginning (B), inside (I) or outside (O) of the answer (see Figure~\ref{fig:qa} for example).
The key difference between our work and \cite{yao:2013:NAACL-HLT} is that \cite{yao:2013:NAACL-HLT} needs a lot work on feature engineering which further relies on POS/NER tagging, dependency parsing, question type analysis, etc. While we avoid feature engineering, and only use one {\em single} model to solve the problem.
Furthermore, compared with sequence generation and classification/ranking methods for answer production, our method avoids expensive $\mathrm{softmax}$ computation and can handle unseen answers/words naturally in a principled way.

Formally, we formalize QA as a sequence labeling problem as follows: suppose we have a vocabulary $V$ of size $|V|$, given question $\mathbf{x^q} = (x_1^q, x_2^q,\cdots,x_N^q)$ and evidence $\mathbf{x^e} = (x_1^e, x_2^e,\cdots,x_M^e)$, where $x_i^q$ and $x_j^e$ are one-hot vectors of dimension $|V|$, and $N$ and $M$ are the number of words in the question and evidence respectively.
The problem is to find the label sequence $\mathbf{\hat{y}}$ which maximizes the conditional probability under parameter $\theta$
\begin{equation}
\mathbf{\hat{y}} = \argmax_{\mathbf{y}}p_\theta(\mathbf{y} | \mathbf{x^q}, \mathbf{x^e}).
\end{equation}
In this work, we model $p_\theta(\mathbf{y} | \mathbf{x^q}, \mathbf{x^e})$ by a neural network composed of LSTMs and CRF.

\section{Recurrent Sequence Labeling Model}
\subsection{Overview}
Figure~\ref{fig:main-figure} shows the structure of our model. The model consists of three components: (1) question LSTM for computing question representation; (2) evidence LSTMs for evidence analysis; and (3) a CRF layer for sequence labeling.
The question LSTM in a form of a single layer LSTM equipped with a single time attention takes the question as input and generates the question representation $r_q$.
The three-layer evidence LSTMs takes the evidence, question representation $r_q$ and optional features as input and produces ``features'' for the CRF layer. 
The CRF layer takes the ``features'' as input and produces the label sequence.
The details will be given in the following sections.

\subsection{Long Short-Term Memory (LSTM)}
\label{sec:model:lstm}
Following~\cite{graves:2013:arxiv}, we define $(s', y') = LSTM(x, s, y)$ as a function mapping its input $x$, previous state $s$ and output $y$ to current state $s'$ and output $y'$:
\begin{eqnarray}
	i & = & \sigma(W_{xi}x + W_{yi}y + W_{si}s + b_i)\\
	f & = & \sigma(W_{xf}x + W_{yf}y + W_{sf}s+b_f)\\
	s' & = & fs+i\sigma(W_{xs}x + W_{ys}y + b_s)\\
	o & = & \sigma(W_{xo}x + W_{yo}y + W_{so}s' + b_o)\\
	y' & = & o \tanh(s')
\end{eqnarray}
where  $W_* \in \mathbb{R}^{H\times H}$ are parameter matrices, $b_*\in \mathbb{R}^{H}$ are biases, $H$ is LSTM layer width, $\sigma$ is the $\mathrm{sigmoid}$ function, $i$, $f$ and $o$  are the input gate, forget gate and output gate respectively.

\subsection{Question LSTM}
\label{sec:model:question-lstm}
The question LSTM consists of a single-layer LSTM~\footnote{Multi-layer LSTMs can be used but no gain was observed.} and a single-time attention model.
The question $\mathbf{x^q} = (x_1^q, x_2^q,\cdots,x_N^q)$ is fed into the LSTM to produce a sequence of vector representations $q_1, q_2, \cdots, q_N$
\begin{equation}
	(s_i^q, q_i) = LSTM(\overline{E}x_i^q, s_{i-1}^q, q_{i-1})
\end{equation}
where $\overline{E}\in \mathbb{R}^{D\times |V|}$ is the embedding matrix and $D$ is word embedding dimension.
Then a weight $\alpha_i$ is computed by the single-time attention model for each $q_i$
\begin{equation}
	\alpha_i = \mathrm{softmax}\left(v_q^{\rm T}\tanh(W_aq_i)\right)
	\label{eqn:weight}
\end{equation}
where $v_q\in\mathbb{R}^D$ and $W_a\in \mathbb{R}^{D\times D}$.
And finally the weighted average $r_q$ of $q_i$ is used as the representation of the question
\begin{equation}
	r_q = \sum_i\alpha_iq_i .
	\label{eqn:weight-avg}
\end{equation}

\subsection{Evidence LSTMs}
\label{sec:model:evidence-lstm}
The three-layer evidence LSTMs processes evidence $\mathbf{x^e} = (x_1^e,$$ x_2^e,\cdots,x_M^e)$ to produce ``features'' for the CRF layer.

The first LSTM layer takes evidence $\mathbf{x^e}$, question representation $r_q$ and optional features as input. We find the following two simple common word indicator features are effective:
\begin{itemize}
	\item {\bf Question-Evidence common word feature (q-e.comm)}: for each word in the evidence, the feature has value 1 when the word also occurs in the question, otherwise 0. The intuition is that words occurring in questions tend not to be part of the answers for factoid questions.
	\vspace{-0.5em}
	\item {\bf Evidence-Evidence common word feature (e-e.comm)}: for each word in the evidence, the feature has value 1 when the word occurs in another evidence, otherwise 0. The intuition is that words shared by two or more evidences are more likely to be part of the answers.
\end{itemize}
Although counterintuitive, we found non-binary e-e.comm feature values does not work well. Because the more evidences we considered, the more words tend to get non-zero feature values, and the less discriminative the feature is.

The second LSTM layer stacks on top of the first LSTM layer, but processes its output in a reverse order. The third LSTM layer stacks upon the first and second LSTM layers with cross layer links, and its output serves as features for CRF layer.

Formally, the computations are defined as follows
\begin{eqnarray}
	x_{e1} & = & [\overline{E}x_j^e;r_q;\overline{F}_1g_j^1;\overline{F}_2g_j^2]\\
	(s_j^{1}, e_j^1) & = & LSTM(x_{e1}, s_{j-1}^{1}, e_{j-1}^1)\\
	(s_j^{2}, e_j^2) & = & LSTM(e_j^1, s_{j+1}^{2}, e_{j+1}^2)\\
	(s_j^{3}, e_j^3) & = & LSTM([e_j^1;e_j^2], s_{j-1}^{3}, e_{j-1}^3)
\end{eqnarray}
where $g_j^1$ and $g_j^2$ are one-hot feature vectors, $\overline{F}_1\in \mathbb{R}^{D_1\times 2}$ and $\overline{F}_2\in \mathbb{R}^{D_2\times 2}$ are embeddings for the features, and $D_1$ and $D_2$ are the feature embedding dimensions. Note that we use the same word embedding matrix $\overline{E}$ as in question LSTM.

\subsection{Sequence Labeling}
\label{sec:model:seq-labeling}
Following~\cite{huang:2015:arxiv,zhou:2015:ACL-IJCNLP}, we use CRF on top of evidence LSTMs for sequence labeling. The probability of a label sequence $\mathbf{y}$ given question $\mathbf{x^q}$ and evidence $\mathbf{x^e}$ is computed as
\begin{equation}
	p_\theta(\mathbf{y} | \mathbf{x^q}, \mathbf{x^e}) \propto \exp(\sum_j\mu[y_{j-1},y_j] + \sum_je_j[y_j])
	\label{eqn:crf}
\end{equation}
where $e_j = W_ee_j^3$, $W_e \in \mathbb{R}^{L\times D}$, $L$ is the number of label types,  $\mu[i,j]$ is the transition weight from label $i$ to $j$, and $e_j[i]$ is the $i$-th value of vector $e_j$.

\begin{table*}
	\centering\footnotesize
	\begin{tabular}{|l|r|r|r|r|r|r|r|r|}
		\hline
		\multirow{3}{*}{\bf Dataset} & \multicolumn{2}{c|}{\multirow{2}{*}{\bf Question}} & \multicolumn{4}{c|}{\bf Annotated Evidence} & 
		\multicolumn{2}{c|}{\multirow{2}{*}{\bf Retrieved Evidence}}\\
		\cline{4-7}
		&  \multicolumn{2}{c|}{} & \multicolumn{2}{c|}{\bf Positive} & \multicolumn{2}{c|}{\bf Negative} & \multicolumn{2}{c|}{}\\
		\cline{2-9}
		& {\bf \#} & {\bf Word \#} & {\bf \#} & {\bf Word \#} & {\bf \#} & {\bf Word \#} & {\bf \#} & {\bf Word \#} \\
		\hline
		Train & 36,145 & 374,500 & 140,897 & 10,757,652 & 122,206 & 14,808,758 & 171,838 & 7,233,543\\
		Validation & 3,018 & 36,666 & 5,412 & 233,911 & / & / & 60,351  & 3,633,540 \\
		Test & 3,024 & 36,815 & 5,445 & 234,258 & / & / & 60,465 & 3,620,391\\
		\hline
	\end{tabular}
	\caption{Statistics of WebQA dataset.}
	\label{tab:webqa}
\end{table*}

\section{Training}
The objective function of our model is
\begin{equation*}
	L_\theta(T)=-\sum_i\log\left(p_\theta(\tilde{\mathbf{y}}_i|\mathbf{x}_i^{\mathbf{q}},\mathbf{x}_i^{\mathbf{e}})\right) + \frac{1}{2}\lambda||\theta||^2
	\vspace{-0.5em}
\end{equation*}
where $\tilde{\mathbf{y}}_i$ is the golden label sequence, and $T = \{(\tilde{\mathbf{y}}_i,\mathbf{x}_i^{\mathbf{q}},\mathbf{x}_i^{\mathbf{e}}\}$ is training set.

We use a minibatch stochastic gradient descent (SGD)~\cite{lecun:1998:IEEE} algorithm with rmsprop~\cite{hinton:2012:rmsprop} to minimize the objective function. The initial learning rate is 0.001, batch size is 120, and $\lambda=0.016$. We also apply dropout~\cite{hinton:2012:arxiv} to the output of all the LSTM layers. The dropout rate is 0.05. All these hyper-parameters are determined empirically via grid search on validation set.

\section{WebQA Dataset}
\label{sec:webqa}

In order to train and evaluate open-domain factoid QA system for real-world questions, we build a new Chinese QA dataset named as WebQA. The dataset consists of tuples of (question, evidences, answer), which is similar to example in Figure~\ref{fig:qa}.
All the questions, evidences and answers are collected from web.
Table~\ref{tab:webqa} shows some statistics of the dataset.

The questions and answers are mainly collected from a large community QA website Baidu Zhidao~\footnote{\url{http://zhidao.baidu.com}} and a small portion are from hand collected web documents.
Therefore, all these questions are indeed {\bf asked by real-world users in daily life instead of under controlled conditions}.
All the questions are of single-entity factoid type, which means (1) each question is a factoid question and (2) its answer involves only one entity (but may have multiple words). The question in Figure~\ref{fig:qa} is a positive example, while the question ``Who are the children of Albert Enistein?'' is a counter example because the answer involves three persons.
The type and correctness of all the question answer pairs are verified by at least two annotators.

All the evidences are retrieved from Internet by using a search engine with questions as queries.
We download web pages returned in the first 3 result pages and take all the text pieces which have no more than 5 sentences and include at least one question word as candidate evidences. As evidence retrieval is beyond the scope of this work, we simply use TF-IDF values to re-rank these candidates.

For each question in the training set, we provide the top 10 ranked evidences to annotate (``Annotated Evidence'' in Table~\ref{tab:webqa}). 
An evidence is annotated as positive if the question can be answered by just reading the evidence without any other prior knowledge, otherwise negative.
Only evidences whose annotations are agreed by at least two annotators are retained.
We also provide trivial negative evidences (``Retrieved Evidence'' in Table~\ref{tab:webqa}), i.e. evidences that do not contain golden standard answers.

For each question in the validation and test sets, we provide one major positive evidence, and maybe an additional positive one to compute features.
Both of them are annotated. Raw retrieved evidences are also provided for evaluation  purpose (``Retrieved Evidence'' in Table~\ref{tab:webqa}).

The dataset will be released on the project page \url{http://idl.baidu.com/WebQA.html}.

\section{Evaluation on WebQA Dataset}
\label{sec:exp-webqa}
\subsection{Baselines}
We compare our model with two sets of baselines:

{\bf MemN2N}~\cite{sukhbaatar:2015:NIPS} is an end-to-end trainable version of memory networks~\cite{weston:2014:arxiv}. It encodes question and evidence with a bag-of-word method and stores the representations of evidences in an external memory. A recurrent attention model is used to retrieve relevant information from the memory to answer the question.

{\bf Attentive and Impatient Readers}~\cite{hermann:2015:arxiv} use bidirectional LSTMs to encode question and evidence, and do classification over a large vocabulary based on these two encodings.
The simpler Attentive Reader uses a similar way as our work to compute attention {\em for the evidence}. And the more complex Impatient Reader computes attention after processing each question word.

The key difference between our model and the two readers is that they produce answer by doing classification over a large vocabulary, which is computationally expensive and has difficulties in handling unseen words. However, as our model uses an end-to-end trainable sequence labeling technique, it avoids both of the two problems by its nature.

\begin{table*}
	\centering\footnotesize
	\begin{tabular}{|l|c|c|c|c|c|c|}
		\hline
		\multirow{2}{*}{\bf System}& \multicolumn{3}{c|}{\bf Validation (Strict)} & \multicolumn{3}{c|}{\bf Test (Strict)}\\
		\cline{2-7}
		& {\bf P} & {\bf R} & {\bf F1} & {\bf P} & {\bf R} & {\bf F1} \\
		\hline
		MemN2N & 52.61 & 52.61 & 52.61 & 50.14 & 50.14 & 50.14 \\
		Attentive Reader & {\bf 65.41} & 65.41 & 65.41 & 62.46 & 62.46 & 62.46 \\
		Impatient Reader & 63.05 & 63.05 & 63.05 & 59.83 & 59.83 & 59.83 \\
		Ours & 64.46 & {\bf 87.62} & {\bf 74.28} & {\bf 63.30} & {\bf 87.70} & {\bf 73.53}\\
		\hline
	\end{tabular}
	\caption{Comparison with baselines on the one-word answer subset of WebQA.}
	\label{tab:baselines}
\end{table*}
\begin{table*}
	\centering\scriptsize
	\begin{tabular}{|l|c|c|c|c|c|c|c|c|c|c|c|c|c|}
		\hline
		\multirow{3}{*}{\bf Model} & \multirow{3}{*}{\bf Noise} & \multicolumn{9}{c|}{\bf Annotated Evidence} & \multicolumn{3}{c|}{\bf Retrieved Evidence}\\
		\cline{3-14}
		& & \multicolumn{3}{c|}{\bf Strict (Val.)} & \multicolumn{3}{c|}{\bf Strict (Test)} & \multicolumn{3}{c|}{\bf Fuzzy (Test)} & \multicolumn{3}{c|}{\bf Fuzzy (Test, Voting)}\\
		\cline{3-14}
		& & {\bf P} & {\bf R} & {\bf F1} & {\bf P} & {\bf R} & {\bf F1} & {\bf P} & {\bf R} & {\bf F1} & {\bf P} & {\bf R} & {\bf F1}\\
		\hline\hline
		Softmax & False & 58.21 & 72.90 & 64.73 & 57.42 & 72.32 & 64.02 & 61.53 & 77.49 & 68.59 & 67.53 & 74.18 & 70.70\\
		Softmax ($k$-1) & False & 58.28 & 71.80 & 64.34 & 58.22& 71.83 & 64.31 & 62.48 & 77.08 & 69.02 & 67.06& 73.47 & 70.11 \\
		CRF & False & 63.19 & {\bf 79.21} & {\bf 70.30} & 61.90 & {\bf 77.33} & 68.76 & 66.00 & {\bf 82.45} & 73.32 & 68.97 & 74.64 & 71.69 \\
		\hline\hline
		Softmax & True & 59.74 & 69.11 & 64.08 & 59.38 & 68.77 & 63.73 & 63.58 & 73.63 & 68.24 & 69.75 & 74.72 & 72.15 \\
		Softmax ($k$-1) & True & 59.84 & 67.51 & 63.44 & 59.76 & 67.61 & 63.44 & 64.02 & 72.44 & 67.97 & 69.11 & 73.93 & 71.44\\
		CRF & True & {\bf 64.42} & 75.84 & 69.67 & {\bf 63.72} & 76.09 & {\bf 69.36} & {\bf 67.53} & 80.63 & {\bf 73.50} & {\bf 72.66} & {\bf 76.83} & {\bf 74.69} \\
		\hline
	\end{tabular}
	\caption{Evaluation results on the entire WebQA dataset. }
	\label{tab:main-results}
\end{table*}

\subsection{Evaluation Method}
\label{sec:exp-webqa:metric}
The performance is measured with precision (P), recall (R) and F1-measure (F1)~\footnote{Measures such MAP and MRR are often also used for evaluating QA system. However, as our model gives only conditional probabilities which are not directly comparable for different answers, we will not include these measures in this work.}
\begin{equation}
	P=\frac{|C|}{|A|},	R=\frac{|C|}{|Q|},F1=\frac{2PR}{P+R}
\end{equation}
where $C$ is the list of correctly answered questions, $A$ is the list of produced answers, and $Q$ is the list of all questions~\footnote{As the baselines will produce exactly one answer for each question, P, R and F1 will be identical for them.}.

As WebQA is collected from web, the same answer may be expressed in different surface forms in the golden standard answer and the evidence, e.g. ``北京 (Beijing)'' v.s. ``北京市 (Beijing province)''.
Therefore, we use two ways to count correctly answered questions, which are referred to as  ``strict'' and ``fuzzy'' in the tables:
\begin{itemize}
	\item {\bf Strict matching}: A question is counted if and only if the produced answer is identical to the golden standard answer;
	\item {\bf Fuzzy matching}: A question is counted if and only if the produced answer is a synonym~\footnote{The synonyms will also be released.} of the golden standard answer;
\end{itemize}
And we also consider two evaluation settings:
\begin{itemize}
	\item {\bf Annotated evidence}: Each question has one major annotated evidence and maybe another annotated evidence for computing q-e.comm and e-e.comm features (Section~\ref{sec:model:evidence-lstm}); 
	\item {\bf Retrieved evidence}: Each question is provided with at most 20 automatically retrieved evidences (see Section~\ref{sec:webqa} for details). All the evidences will be processed by our model independently and answers are voted by frequency to decide the final result. Note that a large amount of the evidences are negative and our model should not produce any answer for them.
\end{itemize}

\subsection{Model Settings}
\label{sec:exp-webqa:setting}
If not specified, the following hyper-parameters will be used in the reset of this section: LSTM layer width $H=64$ (Section~\ref{sec:model:lstm}), word embedding dimension $D=64$ (Section~\ref{sec:model:question-lstm}), feature embedding dimension $D_1=D_2=2$ (Section~\ref{sec:model:question-lstm}). The word embeddings are initialized with pre-trained embeddings using a 5-gram neural language model~\cite{bengio:2003:JMLR} and is fixed during training.

We will show that injecting noise data is important for improving performance on retrieved evidence setting in Section~\ref{sec:exp-webqa:whole-dataset}. 
In the following experiments, 20\% of the training evidences will be negative ones randomly selected on the fly, of which 25\% are annotated negative evidences and 75\% are retrieved trivial negative evidences (Section~\ref{sec:webqa}).
The percentages are determined empirically.
Intuitively, we provide the noise data to teach the model learning to recognize unreliable evidence. 

For each evidence, we will randomly sample another evidence from the rest evidences of the question and compare them to compute the e-e.comm feature (Section~\ref{sec:model:evidence-lstm}). We will develop more powerful models to process multiple evidences in a more principle way in the future.

As the answer for each question in our WebQA dataset only involves one entity (Section~\ref{sec:webqa}), we distinguish label Os before and after the first B in the label sequence explicitly to discourage our model to produce multiple answers for a question. For example, the golden labels for the example evidence in Figure~\ref{fig:qa} will became ``Einstein/O1 married/O1 his/O1 first/O1 wife/O1 Mileva/B Mari{\' c}/I in/O2 1903/O2'', where we use ``O1'' and ``O2'' to denote label Os before and after the first B \footnote{All the words in a negative evidence will get label ``O1''.}. ``Fuzzy matching'' is also used for computing golden standard labels for training set.

For each setting, we will run three trials with different random seeds and report the average performance in the following sections.

\subsection{Comparison with Baselines}
\label{sec:exp-webqa:baselines}
As the baselines can only predict one-word answers, we only do experiments on the one-word answer subset of {WebQA}, i.e. only questions with one-word answers are retained for training, validation and test.
As shown in Table~\ref{tab:baselines}, our model achieves significant higher F1 scores than all the baselines. 

The main reason for the relative low performance of MemN2N is that it uses a bag-of-word method to encode question and evidence such that higher order information like word order is absent to the model. We think its performance can be improved by designing more complex encoding methods~\cite{hill:2015:arxiv} and leave it as a future work.

The Attentive and Impatient Readers only have access to the fixed length representations when doing classification. However, our model has access to the outputs of all the time steps of the evidence LSTMs, and scores the label sequence as a whole. Therefore, our model achieves better performance.

\subsection{Evaluation on the Entire WebQA Dataset}
\label{sec:exp-webqa:whole-dataset}
In this section, we evaluate our model on the entire WebQA dataset. The evaluation results are shown in Table~\ref{tab:main-results}. 
Although producing multi-word answers is harder, our model achieves comparable results with the one-word answer subset (Table~\ref{tab:baselines}), demonstrating that our model is effective for both single-word and multi-word word settings.

``Softmax'' in Table~\ref{tab:main-results} means we replace CRF with $\mathrm{softmax}$, i.e. replace Eq. (\ref{eqn:crf}) with
\begin{equation}
	p_\theta(\mathbf{y} | \mathbf{x^q},\mathbf{x^e}) = \prod_k \mathrm{softmax}(e_k[y_k]) 
	\vspace{-0.5em}
\end{equation}
CRF outperforms $\mathrm{softmax}$ significantly in all cases. The reason is that $\mathrm{softmax}$ predicts each label independently, suggesting that modeling label transition explicitly is essential for improving performance.
A natural choice for modeling label transition in $\mathrm{softmax}$ is to take the last prediction into account as in~\cite{bahdanau:2014:arxiv}. The result is shown in Table~\ref{tab:main-results} as ``Softmax($k$-1)''. However, its performance is only comparable with ``Softmax'' and significantly lower than CRF. The reason is that we can enumerate all possible label sequences implicitly by dynamic programming for CRF during predicting but this is not possible for ``Softmax($k$-1)''~\footnote{We think the performance of ``Softmax($k$-1)'' can be improved by beam search and leave it as a future work.}, which indicates CRF is a better choice. 

``Noise'' in Table~\ref{tab:main-results} means whether we inject noise data or not (Section~\ref{sec:exp-webqa:setting}). 
As all evidences are positive under the annotated evidence setting, the ability for recognizing unreliable evidence will be useless. 
Therefore, the performance of our model with and without noise is comparable under the annotated evidence setting.
However, the ability is important to improve the performance under the retrieved evidence setting because a large amount of the retrieved evidences are negative ones. As a result, we observe significant improvement by injecting noise data for this setting.

\begin{table*}
	\centering\footnotesize
	\begin{tabular}{|l|c|c|c|c|c|c|c|}
		\hline
		\multicolumn{2}{|c|}{\bf Settings} & \multicolumn{3}{c|}{\bf Annotated Evidence} & \multicolumn{3}{c|}{\bf Retrieved Evidence (Voting)}\\
		\hline
		{\bf Initialization} & {\bf Joint Training} & {\bf P} & {\bf R} & {\bf F1} & {\bf P} & {\bf R} & {\bf F1}\\
		\hline
		LM embedding & False & {\bf 67.53} & {\bf 80.63} & {\bf 73.50} & {\bf 72.66} & {\bf 76.83} & {\bf 74.69} \\
		LM embedding & True  & 65.16 & 76.76 & 70.48 & 70.15 & 74.09 & 72.06 \\
		Random & True  & 63.37 & 71.52 & 67.19 & 66.74 & 70.11 & 68.38 \\
		\hline
	\end{tabular}
	\caption{Effect of embedding initialization and training. Only fuzzy matching results are shown.}
	\label{tab:embedding}
\end{table*}

\begin{table*}
	\begin{floatrow}
		\capbtabbox{\centering\footnotesize
			\begin{tabular}{|l|c|c|c|}
				\hline
				\multicolumn{4}{|c|}{\bf Annotated Evidence (Fuzzy)}\\
				\hline
				{\bf Settings} & {\bf P} & {\bf R} & {\bf F1} \\
				\hline
				Both & 67.53 & {\bf 80.63} & {\bf 73.50} \\
				w/o q-e.comm & 64.32 & 69.26 & 66.69 \\
				w/o e-e.comm & {\bf 70.07} & 70.30 & 70.18 \\
				\hline
				\hline
				\multicolumn{4}{|c|}{\bf Retrieved Evidence (Voting, Fuzzy)}\\
				\hline
				{\bf Settings} & {\bf P} & {\bf R} & {\bf F1} \\
				\hline
				Both & {\bf 72.66} & {\bf 76.83} & {\bf 74.69} \\
				w/o q-e.comm & 63.97 & 67.77 & 65.81 \\
				w/o e-e.comm & 71.05 & 75.69 & 73.30 \\
				\hline
			\end{tabular}}{
			\caption{Effect of q-e.comm and e-e.comm features.}
			\label{tab:features}
		}
		\capbtabbox{\centering\footnotesize
			\begin{tabular}{|l|c|c|c|}
				\hline
				\multicolumn{4}{|c|}{\bf Annotated Evidence (Fuzzy)}\\
				\hline
				{\bf Settings} & {\bf P} & {\bf R} & {\bf F1} \\
				\hline
				$\mathrm{attention}$ & 67.53 & {\bf 80.63} & {\bf 73.50} \\
				$\mathrm{max}$ & {\bf 67.80} & 78.84 & 72.90 \\
				$\mathrm{average}$ & 67.30 & 78.03 & 72.27 \\
				\hline
				\hline
				\multicolumn{4}{|c|}{\bf Retrieved Evidence (Voting, Fuzzy)}\\
				\hline
				{\bf Settings} & {\bf P} & {\bf R} & {\bf F1} \\
				\hline
				$\mathrm{attention}$ & {\bf 72.66} & {\bf 76.83} & {\bf 74.69} \\
				$\mathrm{max}$ & 72.08 & 76.34 & 74.15 \\
				$\mathrm{average}$ & 71.31 & 75.41 & 73.30 \\
				\hline
			\end{tabular}}{
			\caption{Effect of question representations.}
			\label{tab:q-sum}
		}
	\end{floatrow}
\end{table*}

\subsection{Effect of Word Embedding}
\label{sec:exp-webqa:embedding}

As stated in Section~\ref{sec:exp-webqa:setting}, the word embedding $\overline{E}$ is initialized with LM embedding and kept fixed in training. We evaluate different initialization and optimization methods in this section. The evaluation results are shown in Table~\ref{tab:embedding}.
The second row shows the results when the embedding is optimized jointly during training. The performance drops significantly. Detailed analysis reveals that the trainable embedding enlarge trainable parameter number and the model gets over fitting easily. The model acts like a context independent entity tagger to some extend, which is not desired. For example, the model will try to find any location name in the evidence when the word ``在哪 (where)'' occurs in the question.
In contrary, pre-trained fixed embedding forces the model to pay more attention to the latent syntactic regularities. And it also carries basic priors such as ``梨 (pear)'' is fruit and ``李世石 (Lee Sedol)'' is a person, thus the model will generalize better to test data with fixed embedding. 
The third row shows the result when the embedding is randomly initialized and jointly optimized. The performance drops significantly further, suggesting that pre-trained embedding indeed carries meaningful priors.

\subsection{Effect of q-e.comm and e-e.comm Features}
As shown in Table~\ref{tab:features}, both the q-e.comm and e-e.comm features are effective, and the q-e.comm feature contributes more to the overall performance. The reason is that the interaction between question and evidence is limited and q-e.comm feature with value 1, i.e. the corresponding word also occurs in the question, is a strong indication that the word may not be part of the answer.

\subsection{Effect of Question Representations}
In this section, we compare the single-time attention method for computing $r_q$ ($\mathrm{attention}$, Eq. (\ref{eqn:weight}, \ref{eqn:weight-avg})) with two widely used options: element-wise max operation 
$\mathrm{max}$: $r_q = \max_iq_i$ and element-wise average operation $\mathrm{average}$: $r_q = \frac{1}{N}\sum_iq_i$.
Intuitively, $\mathrm{attention}$ can distill information in a more flexible way from  \{$q_i$\}, while $\mathrm{average}$ tends to hide the differences between them, and $\max$ lies between $\mathrm{attention}$ and $\mathrm{average}$. The results in Table~\ref{tab:q-sum} suggest that the more flexible and selective the operation is, the better the performance is.

\subsection{Effect of Evidence LSTMs Structures}
We investigate the effect of evidence LSTMs layer number, layer width and cross layer links in this section. The results are shown in Figure~\ref{fig:model-depth-and-width}. For fair comparison, we do not use cross layer links in Figure~\ref{fig:model-depth-and-width} (a) (dotted lines in Figure~\ref{fig:main-figure}), and highlight the results with cross layer links (layer width 64) with circle and square for retrieved and annotated evidence settings respectively. We can conclude that: (1) generally the deeper and wider the model is, the better the performance is; (2) cross layer links are effective as they make the third evidence LSTM layer see information in both directions.

\begin{table*}
	\begin{floatrow}
		\capbtabbox{\centering\footnotesize
			\includegraphics[width=0.8\linewidth]{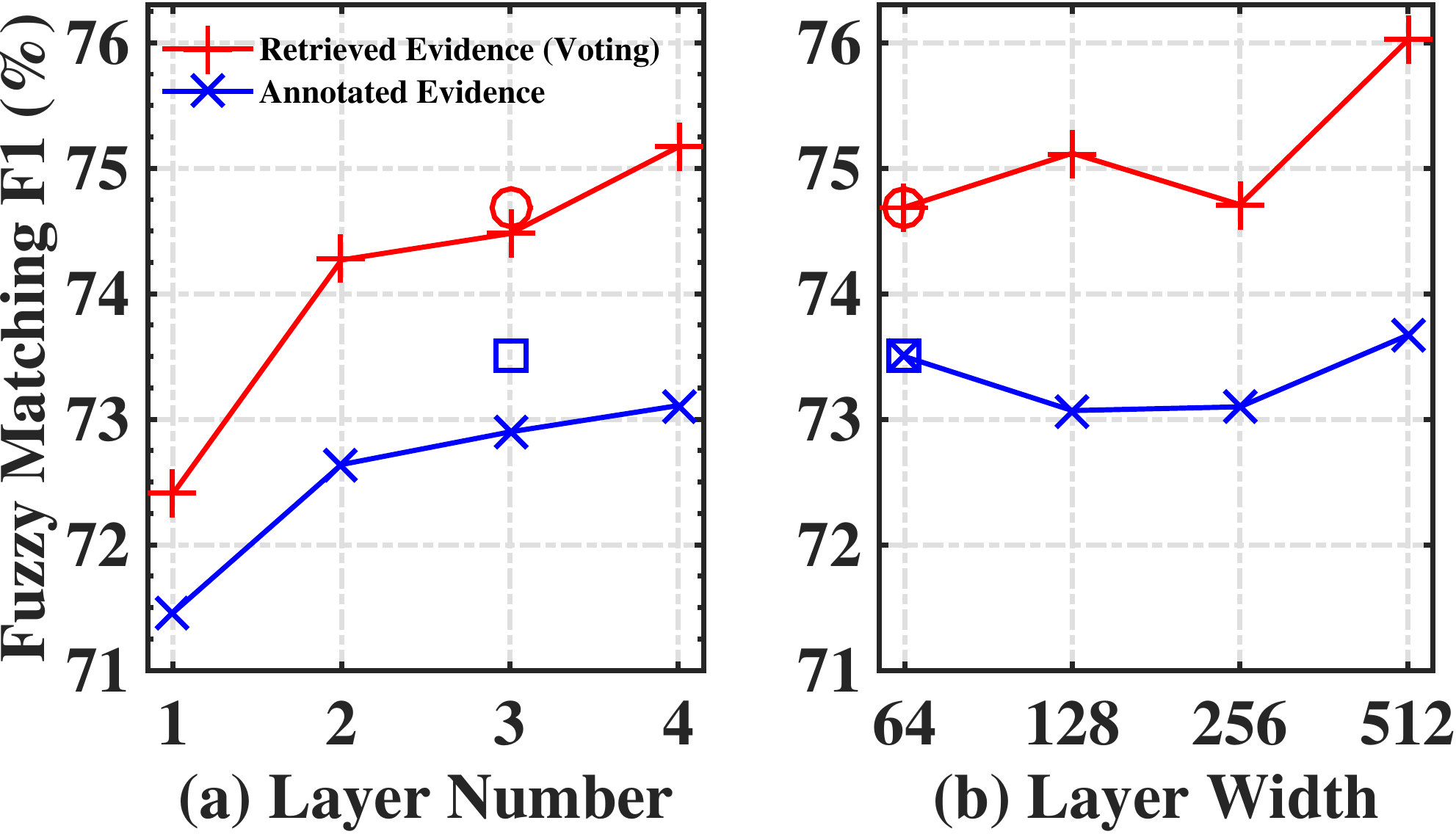}}{
			\figcaption{Effect of evidence LSTMs structures. For fair comparison, cross layer links are not used in (a).}
			\label{fig:model-depth-and-width}
		}
		\capbtabbox{\centering\footnotesize
		\begin{tabular}{|l|c|c|c|}
			\hline
			\multicolumn{4}{|c|}{\bf Annotated Evidence (Fuzzy)}\\
			\hline
			{\bf Model} & {\bf P (\%)} & {\bf R (\%)} & {\bf F1 (\%)} \\
			\hline
			Word-based & {\bf 67.53} & {\bf 80.63} & {\bf 73.50} \\
			Char-based & 67.00 & 72.80 & 69.78 \\
			\hline
			\hline
			\multicolumn{4}{|c|}{\bf Retrieved Evidence (Voting, Fuzzy)}\\
			\hline
			{\bf Model} & {\bf P (\%)} & {\bf R (\%)} & {\bf F1 (\%)} \\
			\hline
			Word-based & {\bf 72.66} & {\bf 76.83} & {\bf 74.69} \\
			Char-based & 68.88 & 73.20 & 70.97 \\
			\hline
		\end{tabular}}{
			\caption{Word-based v.s. character-based input.}
			\label{tab:word-vs-char}
		}
	\end{floatrow}
\end{table*}

\subsection{Word-based v.s. Character-based Input}
Our model achieves fuzzy matching F1 scores of 69.78\% and 70.97\% on character-based input in annotated and retrieved evidence settings respectively (Table~\ref{tab:word-vs-char}), which are only 3.72 and 3.72 points lower than the corresponding scores on word-based input respectively.
The performance is promising, demonstrating that our model is robust and effective.

\section{Conclusion and Future Work}
In this work, we build a new human annotated real-world QA dataset WebQA for developing and evaluating QA system on real-world QA data. We also propose a new end-to-end recurrent sequence labeling model for QA. Experimental results show that our model outperforms baselines significantly.

There are several future directions we plan to pursue. First, multi-entity factoid and non-factoid QA are also interesting topics. Second, we plan to extend our model to multi-evidence cases. Finally, inspired by Residual Network~\cite{he:2015:arxiv}, we will investigate deeper and wider models in the future.

\end{CJK}
	
\bibliography{qa-2016}
\bibliographystyle{emnlp2016}
	
\end{document}